\documentclass[journal,10pt,draftclsnofoot,onecolumn]{IEEEtran}

\usepackage{cite}
\usepackage[cmex10]{amsmath}
\usepackage{algorithmic}
\usepackage{array}
\usepackage{mdwmath}
\usepackage{mdwtab}
\usepackage{eqparbox}
\usepackage[tight,footnotesize]{subfigure}
\usepackage[caption=false,font=footnotesize]{subfig}
\usepackage{url}

\usepackage{amssymb,amsthm}

\usepackage{graphicx}
\usepackage{comment}
\usepackage{algorithm}

\usepackage[english]{babel}
\usepackage[latin5]{inputenc}

\newtheorem{exam}{Example}
\newtheorem{rem}{Remark}

\newcommand{\E}[1]{{E\left[ #1 \right]} }
\newcommand{\partiald}{d}

\newcommand{\Alpha}{$\alpha$-}
\newcommand{\Beta}{$\beta$-}
\newcommand{\AlphaBeta}{$\alpha/\beta$-}
\newcommand{\LL}[2]{\mathcal{L}_{#1}(#2)}

\title{Alpha/Beta Divergences and Tweedie Models}

\author{
Y. Kenan Yılmaz \qquad A. Taylan Cemgil  \\
Department of Computer Engineering\\
Boğaziçi University, Istanbul, Turkey \\
\texttt{kenan@sibnet.com.tr}, \texttt{taylan.cemgil@boun.edu.tr }
}

\begin{document}

\maketitle

\begin{abstract}
We describe the underlying probabilistic interpretation of alpha and beta divergences.  We first show that beta divergences are inherently tied to Tweedie distributions, a particular type of exponential family, known as exponential dispersion models. Starting from the variance function of a Tweedie model, we outline how to get alpha and beta divergences as special cases of Csiszár's $f$ and Bregman divergences. This result directly generalizes the well-known relationship between the Gaussian distribution and least squares estimation to Tweedie models and beta divergence minimization.
 \end{abstract}

\begin{keywords}
  Tweedie distributions, variance functions, alpha/beta divergences, deviance.
\end{keywords}

\section{Introduction}

In fitting a model to data, the error between the model prediction and observed data
can be quantified by a divergence function. The sum-of-squares (Euclidean) cost is an example of such a divergence. It is well know, that minimizing the sum-of-squares error is equivalent to assuming a Gaussian distributed error term and leads to a least squares algorithm. In the recent years, researchers have started using alternative divergences in applications such as KL (Kullback-Leibler) \cite{lee01} or Itakura-Saito (IS) \cite{fevotte09b} divergences. It turns out, that these divergences are special cases of a more general family of divergences known as \Beta divergence \cite{cichocki09}. A different but related family are the \Alpha divergences. Iterative divergence minimization algorithms exist for both families \cite{cichocki09}, however it is often not clear which divergence should be used in an application and it is not clear what the equivalent noise distribution is. In this context, our goal is to survey and investigate results about the relationship between  $\alpha$ and \Beta divergences and their statistical interpretation as a noise model. We believe that it is valuable to have a framework where different divergence functions can be handled without having to invent optimization algorithms from scratch, an aspect of central importance in practical work. We finish the paper by illustrating how the best divergence function can be chosen by maximum likelihood. Moreover, having a deeper understanding of the statistical interpretation of divergence functions could further facilitate model assessment, comparison and improvement.

The motivation of this technical report is i) to  present the central role of the {\it variance functions} in unifying $\alpha$ and \Beta divergences and Tweedie distributions, ii) to provide a compact and simple derivation that unify many results scattered in the statistics and information theory literature about $\alpha$ and \Beta divergences related to Tweedie models. The main observations and contributions of this report are:
\begin{enumerate}
  \item We show that the dual cumulant function of Tweedie distributions  generates $\alpha$ and \Beta divergences.
\item  We simplify and unify the connection between $\alpha$ and \Beta divergences, related scale-invariance properties, the fact that KL divergence is the unique divergence that is both an $\alpha$ and a \Beta divergence and conditions for symmetric \Alpha divergences.
\item  The \Beta divergence is shown to be equivalent to statistical unit deviance, the scaled log-likelihood ratio of a full model to a parametric model.
\item  The density of dispersion models is reformulated using \Beta divergences.
\end{enumerate}
Probability models and divergences are inherently related concepts as shown by various studies; Banerjee et {\it al.} prove the bijection between Bregman divergences and exponential family distributions \cite{banerjee05}. Cichocki et {\it al.}  mention the connection between Tweedie distributions and \Beta divergences in their seminal book \cite{cichocki09}, but very briefly in a single paragraph. Our paper carries their observation one step further by establishing the mathematical formalization based on the the concept of { \it variance functions } \cite{jorgensen1997}. A variance function defines the relationship between the mean and variance of a distribution. For example, the special choice of no functional relationship between the mean and variance (as in linear regression) implies Gaussianity. We show that a power relationship is sufficient to derive \Beta divergences from Bregman divergences and \Alpha divergences from $f$-divergences. This result shows us that using a \Beta divergence in a model is actually equivalent to assuming a Tweedie density.
 For \Alpha divergences, such a direct interpretation is less transparent; but we  illustrate a very direct connection to \Beta divergences and the implicit invariance assumptions about data when \Alpha divergences are used.

\section{Background}

 In this paper, we only consider separable divergences
\begin{align}
  D(\mathbf{x},\mathbf{\mu}) = \sum_i^n d(x_i,\mu_i).
\end{align}
To make the notation simpler we drop the sum from the equations and simply work with scalar divergences $d$. In particular, $d_f(x,\mu)$ denotes $f$-divergence between $x$ and $\mu$ generated by the convex function $f$. Similarly $d_\phi(\cdot,\cdot)$ is the Bregman divergence generated by convex function $\phi$, whereas $d_\alpha(\cdot,\cdot)$ and $d_\beta (\cdot,\cdot)$, simply $\alpha/\beta$,  will denote alpha ($\alpha$) and beta ($\beta$) divergences as special cases. Provided that type of divergence (alpha or beta) is clear from the context, we may replace  alpha, beta symbols with the index parameter $p$ such as in  $d_p(\cdot, \cdot)$. Log-likelihood is denoted by $\LL{x}{\mu}$. In this paper, we assume only univariate case and consider only scalar valued functions whereas the work can easily be extended to multivariate case.

\subsection{Exponential Dispersion Models and Tweedie Distributions}

{\it Exponential Dispersion Models } (EDM) are a linear exponential family defined as \cite{jorgensen87}
\begin{align}
   p(x| \theta,\varphi) = h(x,\varphi)\exp \left\{ \varphi^{-1} \left( \theta x - \psi(\theta) \right)   \right\},
\end{align}
where $\theta$ is the {\it canonical (natural) parameter}, $\varphi$ is the {\it dispersion parameter} and $\psi$ is the cumulant generating function ensuring normalization. Here, $h(x,\varphi)$ is the {\it base measure} and is independent of the canonical parameter. The mean parameter (also called {\it expectation parameter}) is denoted by $\mu$ and is tied to the canonical parameter $\theta$  with the differential equations
\begin{align}
   \mu &= \mu(\theta) =  \frac{\partiald \psi(\theta)}{\partiald\theta}, \label{eq.expectation} \qquad
   \theta = \theta(\mu) = \frac{\partiald \phi(\mu)}{\partiald \mu},
\end{align}
where $\phi(\mu)$ is the conjugate dual of $\psi(\theta)$ just as the canonical parameter $\theta$ is conjugate dual of  expectation parameter $\mu$. The relationship between $\theta$  and  $\mu$ is more direct and  given as \cite{jorgensen87}
\begin{align}
  \frac{ \partiald  \theta}{ \partiald \mu}  = v(\mu)^{-1}.
\end{align}
 Here $v(\mu)$ is the {\it variance function} \cite{tweedie84,barlev86,jorgensen87}, and is related to the variance of the distribution by the dispersion parameter
 \newcommand{\Var}[1]{\mathsf{V}\br{#1}}
  \begin{align}
  Var(x) = \varphi v(\mu).
\end{align}

As a special case of EDMs, {\it Tweedie distributions} $Tw_p(\mu,\varphi)$ specify the variance function as
\begin{align}
   v(\mu) = \mu^{p}
\end{align}
that fully characterizes the dispersion model. The variance function is related to the $p$'th power of the mean, therefore it is called a {\it power variance function} (PVF) \cite{barlev86,jorgensen87}. Here, the special choices of $p=0,1,2,3$ lead to well known distributions as Gaussian, Poisson, gamma and inverse Gaussian. For $1<p<2$, they can be represented as the Poisson sum of gamma distributions so-called {\it compound Poisson distribution}. Indeed, a distribution exist for all real values of $p$ except for $0 < p < 1$ \cite{jorgensen87}.
History of Tweedie distributions goes back to Tweedie's unnoticed work in 1947 \cite{tweedie1947}. Nelder and Wedderburn, in 1972, published a seminal paper on {\it generalized linear models} (GLMs) \cite{nelder1972}, however, without any reference to Tweedie's work where the error distribution formulation was essentially identical to Tweedie's formulation. In 1982, Morris used the term {\it natural exponential models} (NEF) \cite{morris1982}, and 1987 Jorgensen \cite{jorgensen87} coined the name Tweedie distribution.

\subsection{Bregman Divergences and Csiszár $f$-Divergences}

As detailed in the introduction, in many applications it is more convenient to think of minimization of the divergence between data and model prediction. Yet,
probability models and divergences are inherently related concepts \cite{banerjee05}. Two general families of divergences are {\it Bregman divergences} and {\it Csiszár $f$-Divergences}.
Bregman divergences are introduced by  Bregman in 1967. By definition, for any real valued differentiable convex function $\phi$ the Bregman divergence
is given by  \cite{banerjee05}
\begin{align}
   d_\phi(x,\mu) = \phi(x) - \phi(\mu) -(x-\mu) \phi'(\mu).
\end{align}
It is equal to tail of first-order Taylor expansion of $\phi(x)$ at $\mu$. Major class of the cost functions can be generated by the Bregman divergence with appropriate functions $\phi$ as \cite{banerjee05}


\begin{eqnarray*} \label{bregmanB}
     d_\phi(x, \mu)  =
     \left\{
     \begin{array}{ccc} 
     \frac{1}{2} (x-\mu)^2                  & \text{EU} & \phi(x)=\frac{1}{2} x^2  \\
     x \log \frac{x}{\mu} -x  + \mu         & \text{KL} & \phi(x)=x\log x  \\
    \frac{x}{\mu} -\log \frac{x}{\mu}  - 1 & \text{IS} & \phi(x)=-\log x
   \end{array}
     \right.
\end{eqnarray*}

These functions may look arbitrary at a first sight but in the next section we will show that they follow directly from the power variance function assumption.

The {\it f-divergences} are introduced independently by Csiszár \cite{csiszar1963}, Morimoto and Ali \& Silvey during 1960s. They generalize Kullback-Liebler's KL divergences dated back to 1954. By definition, for any real valued convex function $f$ providing that $f(1) = 0$, the $f$-divergence
is given by \cite{csiszar1963}
\begin{align}
   d_f(x,\mu) =  \mu f(\frac{x}{\mu}).
\end{align}
The Bregman and $f$-divergences are non-negative quantities as $d_f(x,\mu) \geq 0$. It is zero iff $x=\mu$, i.e. $d_f(x,x)=0$. Note that the divergences are not distances since they provide neither symmetry nor triangular inequality in general.

\section{Tweedie distributions and Alpha/Beta Divergences}\label{sec3:PowerVarFunc}
In this section, we will derive the link between the Tweedie distributions and \AlphaBeta divergences. We will show that the power variance function assumption is enough to derive both divergences; i.e., if we minimize the \Beta divergence we are assuming a noise density with a power variance function and if we minimize the \Alpha divergence, we assume a certain invariance.

\subsection{Derivation of Conjugate (Dual) of Cumulant Function}
Starting from the power variance assumption, we first obtain the canonical parameter $\theta$ by solving the differential equation $ \frac{\partiald  \theta }{ \partiald \mu } = \mu^{-p}$  \cite{barlev86}
\begin{align}
  \int \partiald\theta  &= \int \mu^{-p} \partiald\mu  \quad \Rightarrow \quad
  \theta = \theta(\mu) = \frac{\mu^{1-p}}{1-p}  + m  \label{eq.theta}
\end{align}
with $m$ is the integration constant. Then we find dual cumulant function $\phi(\cdot)$ by integrating \eqref{eq.expectation} and using $\theta(\mu)$ in \eqref{eq.theta}
\begin{align}
   \phi(\mu) &= \int \theta(\mu) \partiald \mu
   = \int \Big( \frac{\mu^{1-p}}{1-p} + m \Big) \; \partiald \mu  \\
   &=  \frac{\mu^{2-p}}{(1-p)(2-p)}    + m\mu  + d.  \label{edm:dualCumulant}
\end{align}

The $f$-divergence requires $\phi(1)=0$, and for normalization we set $\phi'(1)=0$. Using these two constraints, the constants of integration  are determined as
\begin{align}
   m = -1/(1-p)   \qquad d = 1/(2-p)
\end{align}
so that the dual cumulant function becomes for $p \neq 1,2$
\begin{align}\label{edm:dualCumulant}
     \phi(\mu) &=
     \frac{\mu^{2-p}}{(1-p)(2-p)}  - \frac{\mu}{1-p}  + \frac{1}{2-p}
\end{align}
as the limit cases for $p=1,2$ are found by l'Hôpital's rule


\begin{align*}
     \phi(\mu) &= \left\{
     \begin{array}{ll}
     \frac{1}{2}\mu^2 - \mu +\frac{1}{2}  & \qquad \text{ } p=0  \\
    \mu\log\mu - \mu +1  & \qquad \text{ } p=1  \\
    -\log\mu + \mu -1 & \qquad \text{ } p=2
    \end{array}
    \right.
\end{align*}

The same function $\phi$ is used directly by \cite{lafferty1999, cichocki09} to derive \Beta divergences without justification. Some others \cite{kus2008} obtain it under the name {\it standardized convex form} of the functions by the Bregman divergence as $\phi(\mu)=d_\phi(\mu,1)$.

The function $\phi$ is indeed an entropy function \cite{cichocki09} and can generate a divergence. Similar to \cite{menendez00}, the Shannon's entropy is
\begin{align}
   H[\theta] &= - \int p(x|\theta,\varphi) \log  p(x|\theta,\varphi) \; d\mu(x) \\
    &= -  \varphi^{-1} \big( \theta \mu  - \psi(\theta) \big)  - \E{ \log h(x,\varphi)
   }     \label{edm.entropy}  \\
     H[\mu] &= -  \varphi^{-1}  \phi(\mu)   - \E{ \log h(x,\varphi) }
   \label{edm.entropyMu}
 \end{align}
noting that $\phi(\cdot)$ is the {\it best entropy estimate} where we maximize $\theta \mu  - \psi(\theta)$ to get $H[\mu]$ \cite{borwein1994}.

In the next section, by using the convex function $\phi$, we obtain \Beta divergence from the Bregman divergence and \Alpha divergence from the $f$-divergence.

\subsection{Beta Divergence}

The \Beta divergence is proposed by \cite{eguchi2001, minami2002} and is related to the {\it density power divergence} \cite{basu1998} whereas Cichocki et {\it al.} \cite{cichocki09,fevotte09b} show its relation to Bregman divergence.
Indeed, by use of the dual cumulant function $\phi$, Bregman divergence is specialized to the \Beta divergence
\begin{align}\label{eq:breg.betaDiv0}
   d_\beta(x,\mu) =  \frac{ x^{2-p}}{(1-p)(2-p)}  -\frac{x \mu^{1-p}}{1-p}  + \frac{ \mu^{2-p}}{2-p}
\end{align}
with special cases
\begin{align}\label{eq:breg.betaDiv1}
  d_\beta(x,\mu) &= \left\{
   \begin{array}{ll}
      \frac{1}{2} x^2 -x\mu+ \frac{1}{2}\mu^2 &\quad p=0 \text{ (EU)} \\
      x  \log \frac{x}{\mu} - x + \mu   &\quad p=1 \text{ (KL)}  \\
       - \log \frac{x}{\mu} +\frac{x}{\mu}- 1  &\quad p=2 \text{ (IS)}.   \\
   \end{array} \right.
\end{align}
Note that we can ignore the initial conditions by setting $m=d=0$ in \eqref{edm:dualCumulant},
as for two convex functions $\phi_1,\phi_2$ such that $\phi_1(x)=\phi_2(x) + ax + b$ for some reals $a,b$, $d_{\phi_1}(x,\mu)=d_{\phi_2}(x,\mu)$  \cite{reid2011}; the same divergence is generated if the function $\phi$ is tilted or translated.
The class of distributions induced by the cumulant function $\psi$ are independent of the constants $m$ and $d$ \cite{jorgensen87}. After inverting \eqref{eq.theta} for  solving the parameter $\mu$
\begin{align}\label{muTheta}
    \mu &=  \mu(\theta) =  \left\{  (1-p)  (\theta - m) \right\}^{1/{(1-p)}},
\end{align}
we obtain the cumulant function for the Tweedie distributions as
\begin{align}
  \psi(\theta) &=    \frac{  \left\{ (1-p) (\theta - m) \right\}^{{(2-p)}/{(1-p)}} }{2-p}  + d.  \label{cumulantFn}
\end{align}
Indeed, we can re-parametrize the canonical parameter from $\theta$ to $\theta_1 = \theta - m$  that changes $\psi(\theta_1) = \psi(\theta + m)$ and use $\psi(\theta)$ and $\theta$
 rather than $\psi(\theta_1)$ and $\theta_1$ \cite{barlev86}.

\begin{rem}
The cumulant function parametrized by $\mu$ is  obtained  as
\begin{align}
  \psi(\theta(\mu)) &=  (2-p)^{-1} (\mu^{2-p}-1),  \label{edm.cumulant.mu}
\end{align}
after plugging in $m=-1/(1-p)$ and $d=- 1/(2-p)$ (solve $\psi(0) = 0$ for $d$).  Likewise, the canonical parameter is  $\theta(\mu)=(\mu^{1-p}-1)/(1-p)$ with the limit $\log \mu$ at $p=1$.
\end{rem}

\subsection{Alpha Divergence}

\begin{figure}[!t]
\centering
\subfigure{\includegraphics[trim=0cm 0cm 0cm 7.1cm, clip=true , width=1\columnwidth,height=0.5\columnwidth]{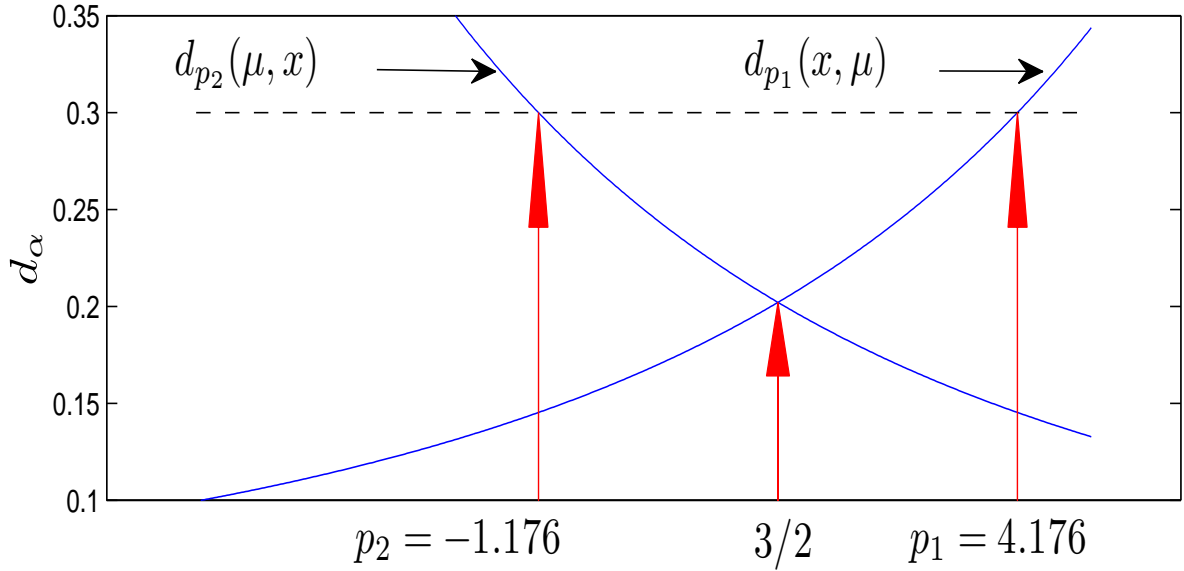}}
\caption{Illustration of the symmetric alpha divergence. Whenever $d_{p_1}(x,\mu)=d_{p_2}(\mu,x)$ (here both are 0.3) the corresponding index values sum to 3, i.e. $p_1+p_2=3$ that consequently  at the point of intersection between two curves is  $p_1=p_2=3/2$.}
\label{fig:alpha1}
\end{figure}

The \Alpha divergence  is a special case of the $f$-divergence \cite{amari1985} obtained by using $\phi$ in \eqref{edm:dualCumulant} as
\begin{align}
   d_\alpha(x,\mu) =  \frac{ x^{2-p} \mu^{p-1}}{(1-p)(2-p)}  -\frac{x}{1-p}  + \frac{\mu}{2-p},
\end{align}
with the special cases
 \begin{align*}
      d_\alpha(x,\mu) = \left\{ \begin{array}{ll}
      \frac{1}{2} \frac{  (x -\mu)^2 }{\mu}           & \text{for } p=0   \\
      x\log \frac{x}{\mu} -x + \mu      & \text{for } p=1   \\
      \mu \log \frac{\mu}{x} +  x - \mu  & \text{for } p=2   \\
      2 \left(x^{1/2}  - \mu^{1/2}  \right)^2 & \text{for } p=3/2. \\
   \end{array}\right.
\end{align*}
 Note the  symmetry for $p=1$ and $p=2$. Here, $p=3/2$ is for the {\it Hellinger distance}, which is a metric satisfying symmetry and the triangular inequality. It is a general rule, in fact, that \Alpha divergences indexed by $p_1,p_2$ enjoy dual relation as illustrated by Figure \ref{fig:alpha1}
\begin{align}
 d_{p_1}(x,\mu) = d_{p_2}(\mu, x) \quad \Leftrightarrow \quad p_1+p_2=3.
\end{align}
The proof is based on the symmetric $f$-divergence $d_f(x,\mu)=d_{f^*}(\mu,x)$, where $f^{*}$ is  {\it Csiszár dual} $f^*(\mu) =\mu f(1/\mu) $ \cite{cichocki11, reid2011}. Note that $d_{3/2}(x,\mu) = d_{3/2}(\mu, x) $, which proves the symmetry for the Hellinger distance.

\begin{rem}
Interestingly setting $p=2-q$ in PVF as ${v(\mu)}  = \mu^{2-q}$ results to
more commonly used form of  \AlphaBeta divergences in the literature \cite{cichocki09, cichocki11, fevotte09b}:
\begin{align}
d_\alpha(x,\mu) &=  {\left(q(q-1)\right)}^{-1} \left\{ x^{q} \mu^{1-q} - qx  + (q-1)\mu \right\}, \\
  d_\beta(x,\mu)  &=
     \left( q(q-1)\right)^{-1} \left\{  x^{q} - (1-q)\mu^{q}   -q x \mu^{q-1} \right\}.
\end{align}
\end{rem}

\subsection{Connection between Alpha and Beta Divergences}

The \Beta divergence can be written as
\begin{align}
    d_\beta(x,\mu) &= {\mu^{1-p}}    \frac{  x^{2-p}\mu^{p-1}  -x(2-p)+\mu(1-p)
 } {(1-p)(2-p)},
\end{align}
where the fraction can be identified as \Alpha divergence. Thus the relation between two divergences is
\begin{align}
  d_\beta(x,\mu) = \mu^{1-p} d_\alpha(x,\mu),
\end{align}
whereas \cite{cichocki2010,cichocki11} give other connections. Note that equality holds for either if $p=1$ or $\mu=1$
\begin{align}
  \mu^{1-p} =1 \qquad \Rightarrow \qquad
   p=1 \text{ or } \mu=1.
\end{align}
The solution for $p=1$, we obtain the KL divergence, the only divergence common to \AlphaBeta divergences \cite{cichocki09}. For $\mu=1$ the divergences are trivially reduced to dual cumulant
\begin{align}
  d_\beta(x,1) = d_\alpha(x,1) = \phi(x).
\end{align}
Here, $\mu=1$ is related to standard uniform distribution \cite{cichocki11} with the entropy of zero  that can be regarded as origin. Also note the relation \cite{zhang2004}
\begin{align}
  d_\alpha(x,\mu) = \mu\phi(\frac{x}{\mu}) =\mu d_\beta(\frac{x}{\mu}, 1).
\end{align}

%

\begin{table}[!t]
\caption{Divergences, distributions and entropies indexed by $p$. The second column is ratio of \Beta and \Alpha divergences. Note also symmetry condition is $p_1+p_2=3$ for two \Alpha divergences.}
\label{table1}
\centering
\begin{tabular}{l|l|l|l|l|l}\hline
{$\mathbf{ p}$} & $\mathbf{\mu^{1-p}}$& {\bf Distribution} & {\bf Beta} & {\bf Alpha} & {\bf Entropy} \\ \hline
$0$ & $\mu$ &  Gaussian & EU & Pearson ($\frac{1}{2}\mathcal{X}^2$) & L$_2$ \\ \hline
$1$ & $1$ & Poisson & KL & KL & Shannon \\ \hline
$\frac{3}{2}$ & $\mu^{-\frac{1}{2}}$ & Comp. Poisson & $-$ & Hellinger dist. &  $-$ \\ \hline
$2$ & $\mu^{-1}$ & Gamma & IS & Reversed KL & Burg \\ \hline
$3$ & $\mu^{-2}$ & Inv. Gaussian & $-$  & Rev. Pearson &  $-$ \\ \hline
\end{tabular}
\end{table}

The findings  are summarized by Table~\ref{table1} that presents the links between \Alpha-type \cite{kus2008} and \Beta-type divergences.

\subsection{Scale Transformation}

Tweeide models $Tw_p(\mu,\varphi)$ are the only dispersion models that provide scale transformation property \cite{jorgensen1997} as  $\forall c \in  \mathbb{R}$
\begin{align}
c \;Tw_p(\mu,\varphi) =   \;Tw_p(c\mu,c^{2-p}\varphi),
\end{align}
whose variance function $v(\mu) = \mu^p$ is scale-invariant since $v(c \mu) =   c^p v(\mu)$. This has corresponding result in divergence side that  \AlphaBeta divergences are scale invariant \cite{cichocki11,fevotte09b}
\begin{align}
  d_\beta(c x, c \mu) &= c^{2-p} d_\beta(x, \mu), \label{scaleInvBeta} \\
  d_\alpha(c x, c \mu) &= c d_\alpha(x, \mu).
\end{align}

\section{Statistical View of Beta Divergence }

\subsection{Density Formulation for Dispersion Models }

Density of EDMs can be re-formulated based on \Beta divergence by using the dual form of  \Beta divergence \cite{banerjee05}
\begin{align}
  d_\beta(x,\mu)  -  \phi(x) &=  - x \theta + \psi(\theta)
\end{align}
that by plugging it in the EDM density, we obtain
\begin{align}
  p(x; \mu,\varphi)
   &= h(x,\varphi) \exp \{\varphi^{-1} (\phi(x) - d_\beta(x,\mu))\}  \\
 &= g(x,\varphi) \exp\{-\varphi^{-1} d_\beta(x,\mu)\}.
\end{align}
Here the base measures $h(\cdot,\cdot)$ and $g(\cdot,\cdot)$ are related as
\begin{align}
  g(x,\varphi) = h(x,\varphi)  \exp\{\varphi^{-1} \phi(x)\}.
\end{align}

\begin{exam}
The Gaussian distribution with dispersion $\varphi=\sigma^2$ can be expressed as a EDM  \cite{jorgensen1997}
\begin{align*}
   p(x;\mu,\sigma^2) &=  \underbrace{(2\pi\sigma^2)^{-\frac{1}{2}} \exp\frac{-x^2}{2\sigma^2}}_{h(x,\varphi)}
     \exp \Big\{\frac{1}{\sigma^2} \Big( x \underbrace{\mu}_{\theta(\mu)} - \underbrace{\frac{\mu^2}{2}}_{\psi(\theta(\mu))}  \Big\}
\end{align*}
the usual form is already expressed as a \Beta divergence
\begin{align*}
   p(x;\mu,\sigma^2) &=  \underbrace{(2\pi\sigma^2)^{-\frac{1}{2}} }_{g(x,\varphi)}
       \exp \Big\{-  \frac{1}{\sigma^2}
       d_\beta(x,\mu) \Big\}
\end{align*}
\end{exam}

\begin{exam}
For the gamma distribution with shape parameter $a$ and inverse scale parameter $b$ , the density is
\begin{align}
  p(x; a,b) = \frac{x^{a-1}}{\Gamma(a)} \exp \{-b x + a \log b\}
\end{align}
The mean and variance of a gamma distribution are given as $\mu=a/b$ and $Var(x) = a/b^2$. Hence, the dispersion becomes $\varphi=1/a$
and we have
$a=1/\varphi$ and $b=1/(\mu\varphi)$. Then the density in terms of mean and dispersion parameters are given as
\begin{align}
  p(x; \mu,a) &= \underbrace{\frac{x^{a-1}}{\Gamma(a)} a^a}_{h(x,\varphi)} \exp \{a ( \underbrace{-\frac{1}{\mu}}_{\theta(\mu)}x  -  \underbrace{\log \mu)}_{\psi(\theta(\mu)) }\}
\end{align}
or equivalently after adding and subtracting $\log x + 1$ from the exponent we obtain
\begin{align}
p(x;\mu,a)
  &= \underbrace{\frac{x^{-1} a^{a} \exp(- a) }{\Gamma(a)}}_{g(x,\varphi)}    \exp \{-a  d_\beta(x,\mu) \}
\end{align}
\end{exam}

\begin{exam}
For the Poisson distribution with dispersion $\varphi=1$, the density is given as \cite{jorgensen1997}
\begin{align}
  p(x;\mu) = \underbrace{\frac{1}{x!}}_{h(x)} \exp \{x \underbrace{\log\mu}_{\theta(\mu)} - \underbrace{\mu}_{\psi(\theta(\mu))}\}
\end{align}
 that after adding and subtracting $x\log x -x$ in the exponent we obtain equivalent beta representation of the density
\begin{align}
  p(x;\mu) =  \underbrace{\frac{x^x \exp x}{x!}}_{g(x)} \exp \{-d_\beta(x,\mu)\}
\end{align}
\end{exam}

This form of density formulation differs from so-called {\it standard form of dispersion models} \cite{jorgensen1997} only the factor $1/2$ in the exponent, that is expressed as
\begin{align}
    p(x; \mu,\varphi) = g(x, \varphi) \exp \{-\frac{1}{2}\varphi^{-1} d_\nu(x, \mu) \},
\end{align}
where $d_\nu$ is {\it unit deviance} ('unit' implies that deviance is scaled by dispersion). The {\it deviance} is a statistical term to qualify the fit of the statistics to the model \cite{nelder1972} and  is equal to  $2$ times of the {\it log-likelihood ratio} of the full model to parametrized model. Log-likelihood of the full model is the maximum achievable likelihood and independent of the parameter $\mu$.
Hence \Beta divergence is linked to the unit deviance and the log-likelihood ratio for given dispersion $\varphi$ as
\begin{align}\label{eq:betaDeviance21}
    d_\beta(x,\mu)  = \frac{1}{2} d_\nu(x,\mu) = \varphi \big\{ \LL{x}{x} -  \LL{x}{\mu} \big\}.
\end{align}

\subsection{Parameter Optimization}

 The last equation \eqref{eq:betaDeviance21} has an immediate result for optimization of expectation parameter $\mu$ for given $p$ and $\varphi$
 \begin{align}
  \frac{\partial d_\beta(x,\mu)}{\partial \mu} = - \varphi \frac{\partial \LL{x}{\mu}}{\partial \mu} = - \frac{(x-\mu)}{\mu^{p}}
\end{align}
that the opposite sign implies minimizing \Beta divergence is equal to the maximizing log-likelihood. Whereas the optimization wrt  $\mu$ is trivial, the likelihood equation
derived by simply plugging $\theta(\mu)$ and $\psi(\theta\mu))$ in log-density of EDM
\begin{align}\label{edm.likelihood}
    \LL{x}{\mu, \varphi, p} =  \varphi  \left\{  \frac{x\mu^{1-p} }{1-p}
 - \frac{\mu^{2-p} }{2-p}  \right\} + \log h(x,\varphi, p)
\end{align}
presents difficulty for optimization of $p$ and  $\varphi$ due to that the base measure $h(x,\varphi, p)$ has no closed forms except for certain values of $p$ as for $p=0,1,2,3$.  For others, such as  $p\in(1,2)$ for compound Poisson the function $h$ is expressed as series \cite{jorgensen1997}. There are a number of approximating techniques one that {\it saddlepoint approximation}, that is  interpreted as being half way between original density and Gaussian approximation as vanishing dispersion $\varphi \to 0$ \cite{jorgensen1997}. Others are Fourier inversion of cumulant generating function and direct series expansion where we refer to \cite{dunn2005} for the details.

\section*{Acknowledgment}

We thank Cedric Fevotte for stimulating discussions on \Beta divergences.

\bibliographystyle{IEEEtran}
\bibliography{Edm23}

\begin{thebibliography}{10}
\providecommand{\url}[1]{#1}
\csname url@samestyle\endcsname
\providecommand{\newblock}{\relax}
\providecommand{\bibinfo}[2]{#2}
\providecommand{\BIBentrySTDinterwordspacing}{\spaceskip=0pt\relax}
\providecommand{\BIBentryALTinterwordstretchfactor}{4}
\providecommand{\BIBentryALTinterwordspacing}{\spaceskip=\fontdimen2\font plus
\BIBentryALTinterwordstretchfactor\fontdimen3\font minus
  \fontdimen4\font\relax}
\providecommand{\BIBforeignlanguage}[2]{{%
\expandafter\ifx\csname l@#1\endcsname\relax
\typeout{** WARNING: IEEEtran.bst: No hyphenation pattern has been}%
\typeout{** loaded for the language `#1'. Using the pattern for}%
\typeout{** the default language instead.}%
\else
\language=\csname l@#1\endcsname
\fi
#2}}
\providecommand{\BIBdecl}{\relax}
\BIBdecl

\bibitem{lee01}
D.~D. Lee and H.~S. Seung, ``Algorithms for non-negative matrix
  factorization,'' in \emph{Neural Information Processing Systems}, vol.~13,
  2001, pp. 556--562.

\bibitem{fevotte09b}
C.~Fevotte, N.~Bertin, and J.~L. Durrieu, ``Nonnegative matrix factorization
  with the itakura-saito divergence. with application to music analysis,''
  \emph{Neural Computation}, vol.~21, 2009.

\bibitem{cichocki09}
A.~Cichocki, R.~Zdunek, A.~H. Phan, and S.~Amari, \emph{Nonnegative Matrix and
  Tensor Factorization}.\hskip 1em plus 0.5em minus 0.4em\relax Wiley, 2009.

\bibitem{banerjee05}
A.~Banerjee, S.~Merugu, I.~S. Dhillon, and J.~Ghosh, ``Clustering with
  {B}regman divergences,'' \emph{JMLR}, vol.~6, pp. 1705--1749, 2005.

\bibitem{jorgensen1997}
B.~Jørgensen, \emph{The Theory of Dispersion Models}.\hskip 1em plus 0.5em
  minus 0.4em\relax Chapman \& Hall/CRC Monographs on Statistics \& Applied
  Probability, 1997.

\bibitem{jorgensen87}
B.~Jorgensen, ``Exponential dispersion models,'' \emph{J. of the Royal
  Statistical Society. Series B}, vol.~49, pp. 127--162, 1987.

\bibitem{tweedie84}
M.~C. Tweedie, ``An index which distinguishes between some important
  exponential families,'' \emph{Statistics: applications and new directions,
  Indian Statist. Inst., Calcutta}, pp. 579--604, 1984.

\bibitem{barlev86}
S.~K. Bar-Lev and P.~Enis, ``Reproducibility and natural exponential families
  with power variance functions,'' \emph{Annals of Stat.}, vol.~14, 1986.

\bibitem{tweedie1947}
M.~C. Tweedie, ``Functions of a statistical variate with given means, with
  special reference to laplacian distributions,'' \emph{Proceedings of the
  Cambridge Philosophical Society}, vol.~49, pp. 41--49, 1947.

\bibitem{nelder1972}
J.~A. Nelder and R.~W.~M. Wedderburn, ``Generalized linear models,'' \emph{J.
  of the Royal Statistical Society, Series A}, vol. 135, pp. 370--384, 1972.

\bibitem{morris1982}
C.~N. Morris, ``Natural exponential families with quadratic variance
  functions,'' \emph{Annals of Statistics}, vol.~10, pp. 65--80, 1982.

\bibitem{csiszar1963}
I.Csiszar, ``Eine informations theoretische ungleichung und ihre anwendung auf
  den beweis der ergodizitt von markoffschen ketten,'' \emph{Publ. Math. Inst.
  Hungar. Acad.}, vol.~8, pp. 85--108, 1963.

\bibitem{lafferty1999}
J.~Lafferty, ``Additive models, boosting, and inference for generalized
  divergences,'' in \emph{Proceedings of Annual Conference on Computational
  Learning Theory}.\hskip 1em plus 0.5em minus 0.4em\relax ACM Press, 1999, pp.
  125--133.

\bibitem{kus2008}
V.~Kus, D.~Morales, and I.~Vajda, ``Extensions of the parametric families of
  divergences used in statistical inference,'' \emph{Kybernetika}, vol.~44,
  no.~1, pp. 95--112, 2008.

\bibitem{menendez00}
M.~L. Menendez, ``Shannon's entropy in exponential families: Statistical
  applications,'' \emph{App. Mathematics Letters}, vol.~13, no.~1, pp. 37--42,
  2000.

\bibitem{borwein1994}
P.~Borwein and A.~Lewis, ``Moment-matching and best entropy estimation,''
  \emph{J of Math. Analysis and App.}, vol. 185, pp. 596--604, 1994.

\bibitem{eguchi2001}
S.~Eguchi and Y.~Kano, ``Robustifing maximum likelihood estimation,'' Institute
  of Statistical Mathematics in Tokyo, Tech. Rep., 2001.

\bibitem{minami2002}
M.~Minami and S.~Eguchi, ``Robust blind source separation by beta divergence,''
  \emph{Neural Computation}, vol.~14, pp. 1859--1886, 2002.

\bibitem{basu1998}
A.~Basu, I.~R. Harris, N.~L. Hjort, and M.~C. Jones, ``Robust and efficient
  estimation by minimising a density power divergence,'' \emph{Biometrika},
  vol.~85, no.~3, pp. 549--559, 1998.

\bibitem{reid2011}
M.~D. Reid and R.~C. Williamson, ``Information, divergence and risk for binary
  experiments,'' \emph{JMLR}, vol.~12, pp. 731--817, 2011.

\bibitem{amari1985}
S.~Amari, \emph{Differential-Geometrical Methods in Statistics}, Editor,
  Ed.\hskip 1em plus 0.5em minus 0.4em\relax Springer Verlag, 1985.

\bibitem{cichocki11}
A.~Cichocki, S.~Cruces, and S.~Amari, ``Generalized alpha-beta divergences and
  their application to robust nonnegative matrix factorization,''
  \emph{Entropy}, vol.~13, no.~1, pp. 134--170, 2011.

\bibitem{cichocki2010}
A.~Cichocki and S.~Amari, ``Families of alpha-beta-and gamma-divergences:
  Flexible and robust measures of similarities,'' \emph{Entropy}, vol.~12, pp.
  1532--1568, 2010.

\bibitem{zhang2004}
J.~Zhang, ``Divergence function, duality, and convex analysis,'' \emph{Neural
  Computation}, vol.~16, pp. 159--195, 2004.

\bibitem{dunn2005}
P.~K. Dunn and G.~S. Smyth, ``Series evaluation of tweedie exponential
  dispersion model densities,'' \emph{Stats. \& Comp.}, vol.~15, pp. 267--280,
  2005.

\end{thebibliography}

\end{document}